%% file: main.tex
\documentclass[10pt,twocolumn,letterpaper]{article}

\usepackage[pagenumbers]{iccv} 

\input{preamble}

\definecolor{iccvblue}{rgb}{0.21,0.49,0.74}
\usepackage[pagebackref,breaklinks,colorlinks,allcolors=iccvblue]{hyperref}

\def\paperID{5} 
\def\confName{ICCV}
\def\confYear{2025}

\title{Smart Eyes for Silent Threats: VLMs and In-Context Learning for THz Imaging}

\author{Nicolas Poggi$^{1}$
\quad
Shashank Agnihotri$^{1}$
\quad
Margret Keuper$^{1,2}$\\
$^{1}$Data and Web Science Group, University of Mannheim, Germany \\
$^{2}$Max-Planck-Institute for Informatics, Saarland Informatics Campus, Germany \\
{\tt\small shashank.agnihotri@uni-mannheim.de}
}

\begin{document}
\maketitle
\begin{abstract}
Terahertz (THz) imaging enables non-invasive analysis for applications such as security screening and material classification, but effective image classification remains challenging due to limited annotations, low resolution, and visual ambiguity. We introduce In-Context Learning (ICL) with Vision-Language Models (VLMs) as a flexible, interpretable alternative that requires no fine-tuning. Using a modality-aligned prompting framework, we adapt two open-weight VLMs to the THz domain and evaluate them under zero-shot and one-shot settings. Our results show that ICL improves classification and interpretability in low-data regimes. This is the first application of ICL-enhanced VLMs to THz imaging, offering a promising direction for resource-constrained scientific domains. Code: 
\href{https://github.com/Nicolas-Poggi/Project_THz_Classification/tree/main}{GitHub repository}.
\end{abstract}

\section{Background}
Terahertz (THz) imaging is a promising modality for non-destructive analysis in security screening, industrial inspection, and material characterization \cite{seferyan2025}. Its ability to penetrate non-metallic materials and capture spectral information enables safe, non-ionizing detection of concealed objects \cite{shen2024transformerbasedmethodregistration, palka2012THzspectroscopyimaging} and material differentiation \cite{kubiczek2022MaterialClassificationTerahertz}. However, effective and interpretable classification of THz images remains challenging.

Traditional machine learning methods are limited by the scarcity of labeled data \cite{shen2024transformerbasedmethodregistration}, high visual ambiguity across materials \cite{kubiczek2022MaterialClassificationTerahertz}, and the low spatial resolution typical of THz systems \cite{seferyan2025}. These factors reduce the generalizability of supervised models and hinder their deployment in safety-critical tasks, where interpretability is essential.

To address these limitations, we investigate In-Context Learning (ICL) \cite{dong2024SurveyIncontextLearning} with Vision-Language Models (VLMs) \cite{li2025SurveyStateArt} as a lightweight and interpretable alternative that adapts to new tasks at inference time using only a few labeled examples. ICL has proven effective in domains such as medical pathology \cite{ferber2024Incontextlearningenables}, marine biology \cite{sun2021FewShotLearningDomainSpecific}, and legal document review \cite{huber-fliflet2024ExperimentalStudyInContext}, but has not been explored in THz imaging.

While proprietary models like Gemini \cite{team2023gemini} and GPT-4o \cite{achiam2023gpt4} offer larger context windows and stronger performance, they are often inaccessible due to cost and API constraints. We focus on open-source VLMs, which are transparent, reproducible, and suitable for offline use. Though limited by shorter context windows, our framework can be extended to more capable models in future work.

This research is motivated by real-world constraints in developing countries, where access to skilled experts for material classification or security screening may be limited. In such settings, VLMs with ICL can serve as cost-effective decision support tools, extending human expertise and improving system transparency.

This work makes the following contributions:
\begin{itemize}
\item We propose a modality-aligned ICL framework for adapting general-purpose VLMs to THz image classification without fine-tuning.
\item We evaluate two open-weight VLMs under zero-shot and one-shot settings, showing performance gains in low-data conditions.
\item We demonstrate that ICL improves interpretability by producing natural language justifications for predictions.
\item We discuss the potential of ICL-based VLMs for deployment in resource-constrained environments.
\end{itemize}

\section{Methodology}

\input{latex_for_figures/THz-Data-Example}

This section outlines the THz imaging setup, dataset construction, preprocessing for visualization, and the use of Vision-Language Models (VLMs) under zero-shot and one-shot In-Context Learning (ICL) settings for frame-wise classification.

\subsection{THz Imaging Setup and Dataset}

The dataset consists of 1,400 frames captured using a focused THz emitter-receiver system designed for non-invasive detection. The setup scans a scene with two objects: a small sphere of C4 explosive encased in a metallic shell, and a nearby metallic plate placed on styrofoam.

Each frame corresponds to a unique THz frequency, spanning a range of 1,400 values chosen to vary penetration depth. Collectively, these frames produce a quasi-3D scan of the objects. The raw sensor data is processed using a Fourier Transform to extract both intensity and phase components, which are visualized as 2D heatmaps. Mid-range frequencies (approximately index 300 to 1200), particularly around frame 700, show the C4 signature most clearly, for example, \Cref{fig:THz-Data-Example}. The created dataset is available \href{https://data.dws.informatik.uni-mannheim.de/machinelearning/THz_VLM_ICL_Classification/Videos.zip}{here}.

\subsection{Data Representation and Annotation}

Each frame consists of dual plots for intensity and phase, along with the index of the capturing frequency. A human annotator labeled all frames as \textit{Yes C4} or \textit{No C4} based on visible spectral features and informed by early zero-shot predictions. These annotations are used as the ground truth for evaluating model outputs, similar to \cite{prasse2023towards}.

\subsection{Zero-Shot Classification with VLMs}

\input{latex_for_figures/THz-Data-Image-To-Crop}
\input{latex_for_figures/Method-InOut-THz-ICL-VLM}

In the zero-shot setting, each frame is independently passed to a VLM along with a textual prompt describing the classification task, including how to interpret THz plots and what C4-related features to expect. The model receives the full dual-plot image and returns a natural language prediction indicating whether C4 is present. No fine-tuning or demonstration examples are used. We evaluate predictions on all 1,400 frames against human annotations.

\subsection{One-Shot In-Context Learning with VLMs}

To test ICL, we adopt a one-shot setup where a single in-context example is prepended before each query. This example consists of a cropped region from a frame where C4 is clearly visible (typically from the mid-frequency range), with frequency metadata and surrounding noise removed.

\begin{itemize}
\item A 26×26 pixel crop is extracted from the region containing the C4 signal.
\item The crop is paired with a short natural language description stating that it contains C4.
\end{itemize}

This visual-textual pair is added before each test frame during inference. The model is then asked whether the query image also contains C4. The crop selection and preprocessing are shown in \Cref{fig:THz-Data-Image-To-Crop}, and the overall pipeline is illustrated in \Cref{fig:InOut-THz-ICL-VLM}.

\subsection{Modality-Aligned Prompting Framework}

We follow a modality-aligned ICL format that maintains positional consistency between visual and textual inputs. The text includes the task instruction, optional demonstration (in one-shot), and the query. The visual input includes the demonstration crop (if applicable) followed by the query image.
This structure encourages the VLM to associate the in-context example with its label and apply that knowledge when classifying the new input.

\subsection{Evaluation Protocol}

We compare model outputs in both zero-shot and one-shot configurations against human annotations. Evaluation metrics include accuracy, precision, recall, and F1-score, allowing us to quantify overall performance and trade-offs between sensitivity and specificity. Details on the metrics are provided in \Cref{sec:appendix:metrics_used}.

\section{Results and Analysis}
\label{sec:results}
\input{latex_for_figures/table-result-standard}
\input{latex_for_figures/table-result-impact-ICL}
\input{latex_for_figures/Qualitative-InOut-THz-ICL-VLM-Models-Example}

We evaluate the effect of In-Context Learning (ICL) on the performance of two Vision-Language Models (VLMs): \texttt{Mistral-Small-3.1-24B-Instruct-2503} and \texttt{Qwen2.5-VL-7B-Instruct}, applied to frame-wise THz image classification. Results are presented in terms of standard classification metrics, prediction behavior changes, and qualitative output analysis across zero-shot and one-shot settings. \Cref{sec:appendix:exp_setup} provides implementation details.

\subsection{Quantitative Evaluation and ICL Impact}

\Cref{tab:experiment-result-standard} summarizes the classification performance of both models. For \textbf{Mistral}, ICL improves overall performance. Accuracy increases from 0.4950 to 0.7193, and the F1-score rises from 0.3825 to 0.4126. While recall decreases slightly, precision improves, indicating that Mistral becomes more conservative but more accurate in its positive predictions.

For \textbf{Qwen}, the effect of ICL is mixed. Accuracy drops from 0.7207 to 0.5329 in the one-shot setting, but recall improves significantly from 0.5000 to 0.9661. This suggests a shift toward over-detection, capturing more true positives at the cost of increased false positives. The F1-score remains relatively stable, indicating a trade-off between sensitivity and specificity rather than a net gain in reliability.

\Cref{tab:experiment-result-impact-ICL} provides additional insight into frame-level prediction dynamics. \textbf{Mistral} improves on 408 frames and declines on only 94, reflecting a positive and stable transition with ICL. In contrast, \textbf{Qwen} improves on 131 frames but degrades on 394, highlighting significant instability when conditioned on a single example. These results suggest that ICL consistently enhances Mistral’s performance, while introducing variability and over-sensitivity in Qwen.

\subsection{Quantitative Evaluation and ICL Impact}

\Cref{tab:experiment-result-standard} summarizes the classification performance of both models across zero-shot and one-shot settings. For \textbf{Mistral}, ICL leads to clear gains: accuracy improves from 0.4950 to 0.7193, and the F1-score increases from 0.3825 to 0.4126. Although recall drops slightly, precision improves, indicating that Mistral becomes more conservative but more accurate in its positive predictions.
In contrast, \textbf{Qwen} shows a mixed response to ICL. Accuracy drops from 0.7207 to 0.5329, while recall increases from 0.5000 to 0.9661, suggesting a shift toward over-detection with more true positives and false positives. The F1-score remains nearly constant, reflecting a trade-off between sensitivity and specificity rather than an overall improvement.
To gain finer-grained insight, \Cref{tab:experiment-result-impact-ICL} reports frame-level prediction dynamics. \textbf{Mistral} improves on 408 frames and declines on only 94, showing a net positive and stable transition. \textbf{Qwen}, however, improves on 131 but declines on 394 frames, highlighting greater prediction volatility. Thus, ICL enhances Mistral’s performance more reliably, while introducing inconsistency and over-sensitivity in Qwen.

\section{Conclusion}
This work presents the first application of In-Context Learning (ICL) with Vision-Language Models (VLMs) for Terahertz (THz) image classification, a domain characterized by scarce annotations, low resolution, and high visual ambiguity. We propose a modality-aligned prompting framework and show that ICL can improve classification performance, particularly for the Mistral model, while revealing model-specific trade-offs in precision and recall. Although overall accuracy remains limited and the system is not yet reliable for deployment, this study provides a foundation for future improvements using fine-tuned or proprietary models. Our results suggest that general-purpose VLMs can be adapted to complex sensing tasks with minimal supervision, as also demonstrated in domains like ultrasound imaging \cite{le2025U2BENCHBenchmarkingLarge}. This direction holds promise for cost-effective, interpretable THz imaging in real-world scenarios such as security screening and material inspection \cite{shen2024transformerbasedmethodregistration, palka2012THzspectroscopyimaging}, particularly in expert-scarce or resource-limited environments.

\noindent\textbf{Acknowledgement. }S.A. and M.K. acknowledge support by the DFG Research Unit 5336 - Learning2Sense (L2S) and would like to thank our L2S collaborators, especially Prof. Dr. Peter Haring and his group for the data, and Prof. Dr. Michael Möller and Prof. Dr. Onofre Martorell for their help with the scripts to interpret the THz data. The authors acknowledge support by the state of Baden-Württemberg through bwHPC.

{
    \small
    \bibliographystyle{ieeenat_fullname}
    \bibliography{main}
}

\input{appendix}

\end{document}

%% file: preamble.tex
\usepackage{algorithm}
\usepackage{algpseudocode}


%% file: latex_for_figures/THz-Data-Example.tex
\begin{figure}[htbp]
    \centering
    \includegraphics[width=\linewidth]{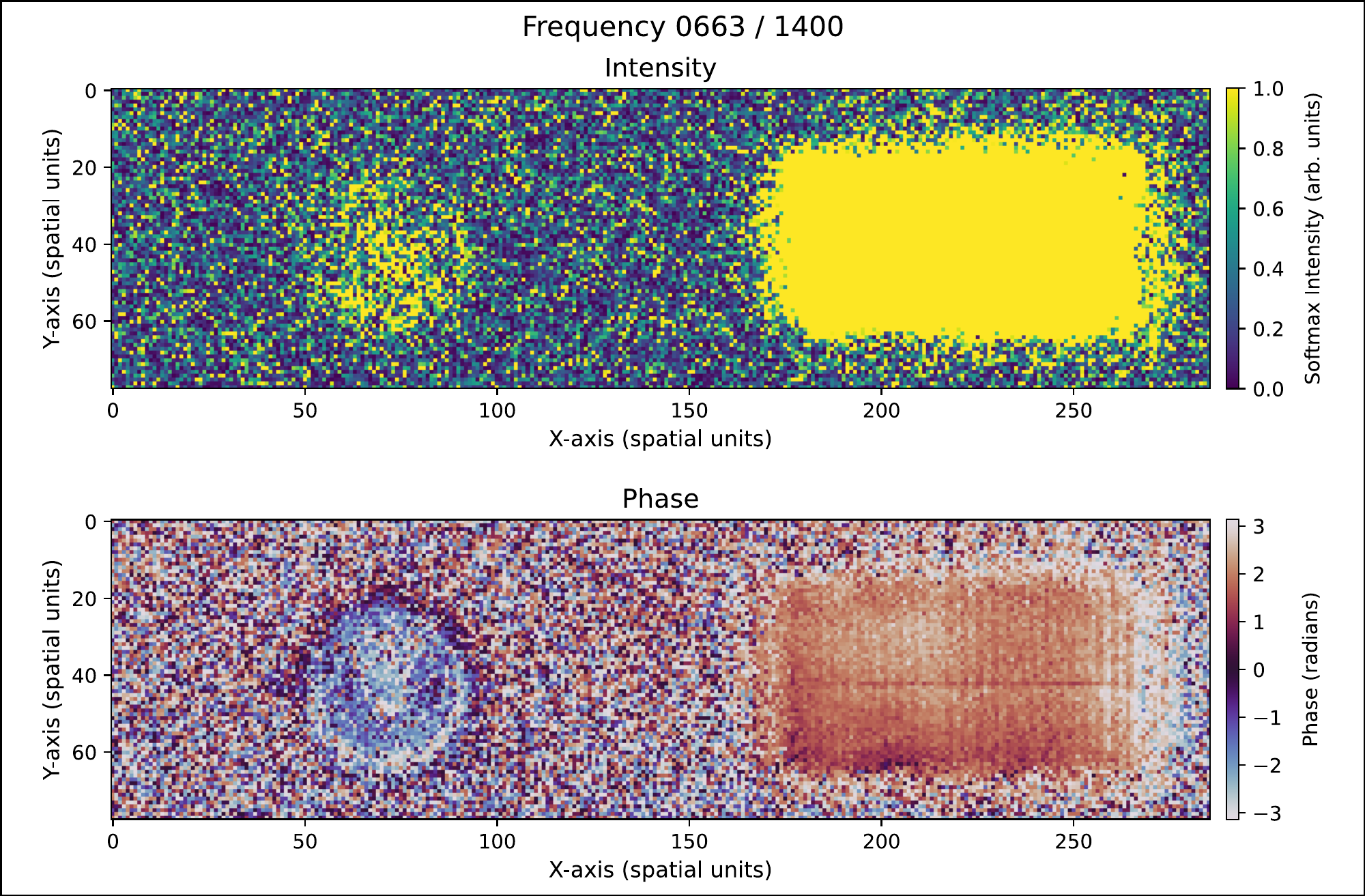}
    \caption{Example frame from the dataset (Frame Number 0663), selected from the total of 1,400 frames. This frame displays intensity (top) and phase (bottom) plots. Each plot shows a circular region on the left indicating the C4 explosive and a rectangular region on the right indicating the metal plate. While the actual experiment displays intensity and phase images side by side (intensity left, phase right), this figure arranges them vertically for better readability.}
    \label{fig:THz-Data-Example}
\end{figure}

%% file: latex_for_figures/THz-Data-Image-To-Crop.tex
\begin{figure}[htbp]
    \centering
    \includegraphics[width=\linewidth]{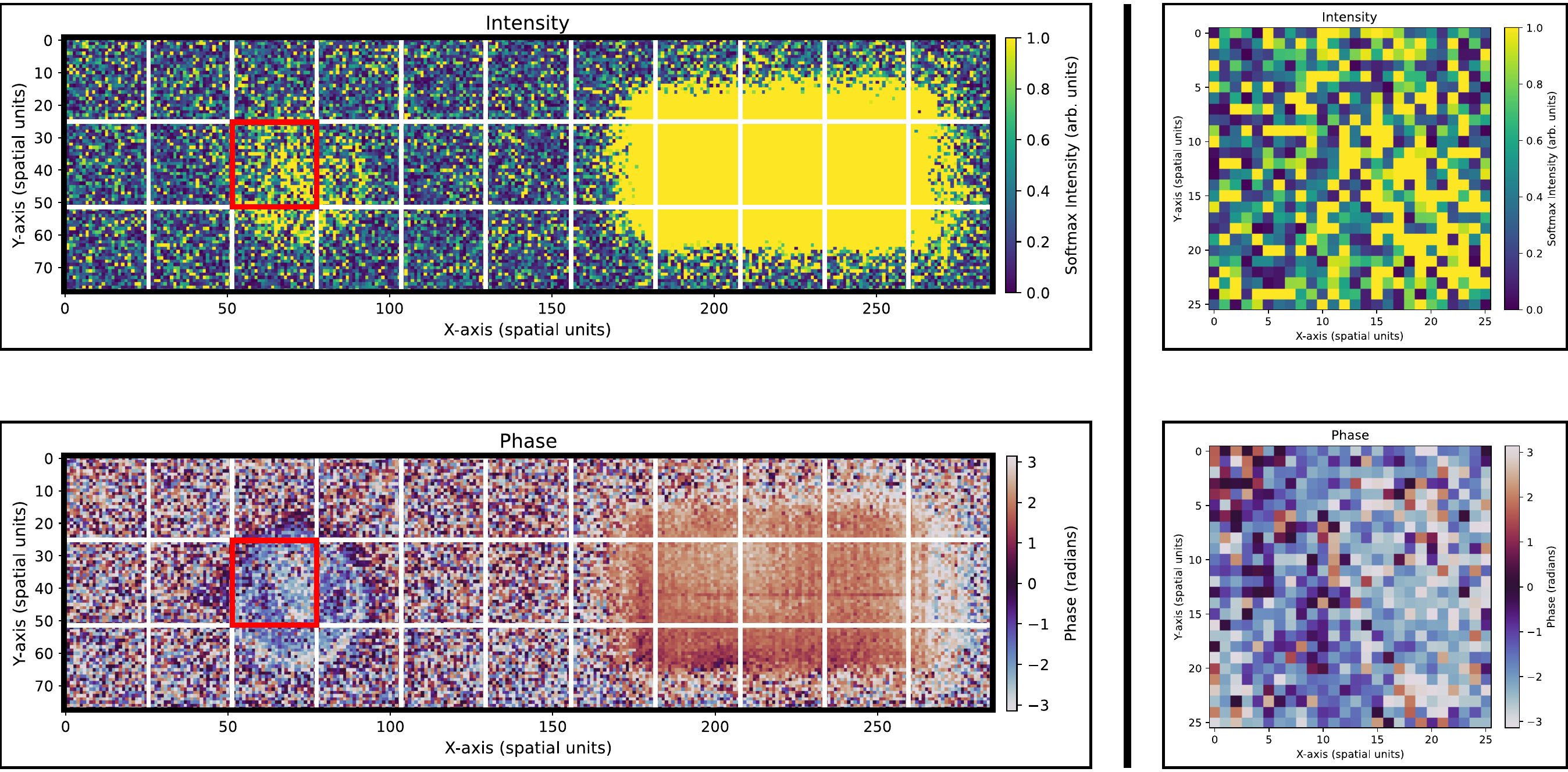}
    \caption{Example illustrating how crops were extracted from the 1,400 frames in the dataset. On the left is the complete frame (Frame Number 0663, also shown in Figure \ref{fig:THz-Data-Example}) displaying intensity and phase plots with the visible “C4” circle. On the right is an enlarged view of a selected crop from this frame.}
    \label{fig:THz-Data-Image-To-Crop}
\end{figure}

%% file: latex_for_figures/Method-InOut-THz-ICL-VLM.tex
\begin{figure*}[h]
    \centering
    \includegraphics[width=\linewidth]{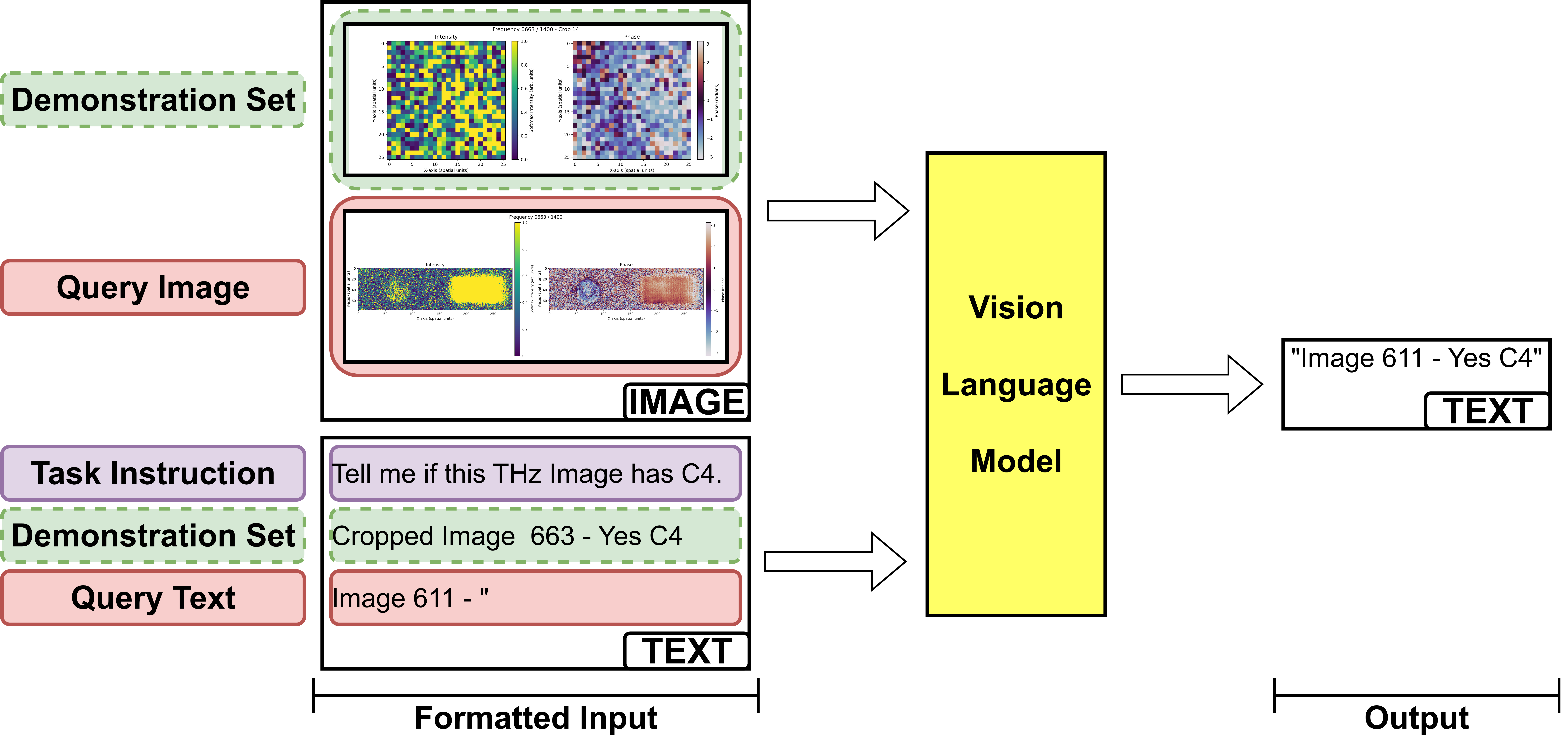}
    \caption{A high-level overview of our In-Context Learning implementation using a Vision Language Model for the Terahertz (THz) image classification task. When the Demonstration Set (both image and text input) is excluded, the model operates in a zero-shot setting. When the Demonstration Set is included, the model performs in a few-shot (one-shot) format. In the actual experiment, the textual input used is longer and more detailed (See Appendix).}
    \label{fig:InOut-THz-ICL-VLM}
\end{figure*}

%% file: latex_for_figures/table-result-standard.tex
\begin{table}
    \centering
    \caption{Quantitative results of two Vision-Language Models (Mistral and Qwen) evaluated on the THz classification task under 0-shot and 1-shot settings. Metrics include accuracy, precision, recall, and F1-Score, illustrating the impact of In-Context Learning (ICL) on model performance.}
    \resizebox{\linewidth}{!}{
        \begin{tabular}{cccccc}\toprule
            \textbf{Model} & \textbf{Shot Format} & \textbf{Accuracy} & \textbf{Precision} & \textbf{Recall} & \textbf{F1-Score} \\\midrule
            Mistral-Small-3.1-24B-Instruct-2503 & 0-Shot & 0.4950 & 0.2409 & 0.9280 & 0.3825 \\
            Qwen2.5-VL-7B-Instruct & 0-Shot & 0.7207 & 0.3018 & 0.5000 & 0.3764 \\\midrule
            Mistral-Small-3.1-24B-Instruct-2503 & 1-Shot & 0.7193 & 0.3187 & 0.5847 & 0.4126 \\
            Qwen2.5-VL-7B-Instruct & 1-Shot & 0.5329 & 0.2609 & 0.9661 &0.4108 \\\bottomrule
        \end{tabular}
    }\label{tab:experiment-result-standard}
\end{table}

%% file: latex_for_figures/table-result-impact-ICL.tex
\begin{table}
    \centering
     \caption{Prediction Changes Metric of VLMs Mistral-Small-3.1-24B-Instruct-2503 and  Qwen2.5-VL-7B-Instruct, with the difference being from the baseline (0-shot) to the comparison (1-shot) experiment.}
    \resizebox{\linewidth}{!}{
    \begin{tabular}{ccccc}\toprule
         Model & Improvement &  Decline & No Improvement & No Decline \\\midrule
         Mistral-Small-3.1-24B-Instruct-2503 & 408 & 94 & 299 & 599 \\
         Qwen2.5-VL-7B-Instruct & 131 & 394 & 260 & 615 \\\bottomrule
    \end{tabular}
    }\label{tab:experiment-result-impact-ICL}
\end{table}

%% file: latex_for_figures/Qualitative-InOut-THz-ICL-VLM-Models-Example.tex
\begin{figure*}[htbp]
    \centering
    \includegraphics[width=0.94\linewidth]{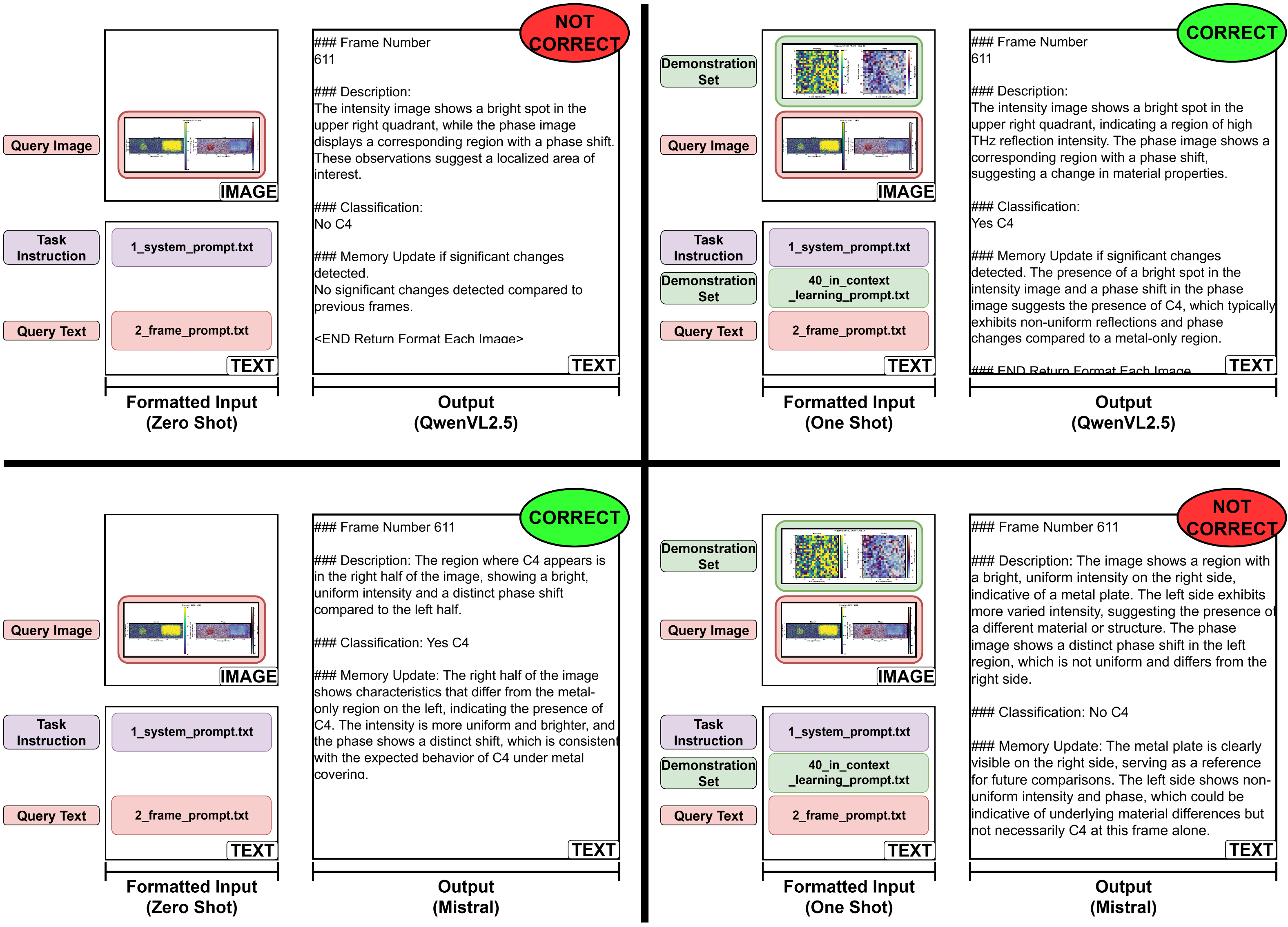}
    \caption{This figure shows example outputs from two Vision Language Models (VLMs): Qwen2.5-VL-7B-Instruct (top) and Mistral-Small-3.1-24B-Instruct-2503 (bottom). For each model, outputs are presented for both zero-shot (left) and one-shot (right) input formats. All settings use the same task description and query prompt templates (see appendix for details), with Frame 611 as the query image. In the one-shot format, an additional demonstration example (consisting of a cropped image from Frame 663 and its corresponding prompt) is included. The correct classification for Frame 611 is ``YES C4". A ``CORRECT” or ``NOT CORRECT” label is displayed in the top-right corner of each output to reflect whether the model’s prediction was accurate.}
    \label{fig:InOut-THz-ICL-VLM-Models-Example}
\end{figure*}

%% file: appendix.tex
\newpage
\appendix

\section{Experimental Setup}
\label{sec:appendix:exp_setup}
\subsection{Models and Prompt Design}
We tested two open-source VLMs from Hugging Face: \texttt{Mistral-Small-3.1-24B-Instruct-2503} and \texttt{Qwen2.5-VL-7B-Instruct}. Each model received a multimodal prompt inlcuding a task instruction, and (depending on the setting) an optional demonstration example followed by a query. Few-shot prompts used a single in-context crop labeled “Yes C4” selected from a random frame. The used prompt templates in the appendix.

\subsection{Environment}
All experiments were run on an AMD Instinct MI300 GPU using the BWUniCluster 3.0 infrastructure. To ensure consistency and reproducibility, we used an Anaconda environment to manage library versions.

\section{Dataset Used}
The dataset used for this work is available as a zip file with downloadable videos in the MP4 format here: \url{https://data.dws.informatik.uni-mannheim.de/machinelearning/THz_VLM_ICL_Classification/Videos.zip}.
One can download the videos and then extract each frame from them, if required.
The data itself was captured and shared by the ``Lehrstuhl Höchstfrequenztechnik und Quantenelektronik'' at the University of Siegen, collaborators under the L2S Research Unit 5336. 
We are thankful for the data. The THz data itself is the property of the ``Lehrstuhl Höchstfrequenztecnik und Quantenelektronik'', University of Siegen and would be released at their discretion.

\section{Prompts Used}
This section presents the prompt templates employed in our experiment. Each classification setup used a combination of the following: a Task Instruction, a Demonstration Set (in the one-shot setting), and a Query prompt. 

\subsection{Full Prompts}
The complete contents of the prompt templates are listed below for reference.
\\\\
\textbf{Query Prompt:}
\begin{quote}
\textit{``Sending the next Image now. Analyse it and answer with reasoning. Here is the Image Nr. [Image NR]''}
\footnotesize\texttt{(2 frame prompt.txt)}
\end{quote}

\noindent\textbf{Demonstration Set Prompt:}
\begin{quote}
\textit{``Use this Image as a reference for Classifying. It has the Class ``YES C4"''}
\footnotesize\texttt{(40 in context learning prompt.txt)}
\end{quote}
\textbf{Task Instruction:}
\begin{quote}
\textit{``- Goal\\
Identify the presence of C4 explosive material by analyzing a sequence of 1400 terahertz (THz) imaging plots. Each image represents a frequency-specific intensity and phase profile across depth. Determine which frequencies (images) and regions most confidently indicate the presence of C4, based on differences in intensity and phase behavior between metal-only and metal+C4 regions. Use the ``Context Dump" Section as a reference for understanding the given THz images and how to be classifying them.
\\\\
In the following prompts I will send each Terahertz image one at a time in the described format (See Prompt Format-Each Image). Please Analyse each Image and Classify it based on $<$Return Format Each Image$>$ and $<$Warnings$>$. The output of the prompt should not repeat the instructions back to me. It should only give a brief answer (``Yes C4" or ``No C4") and answer with reasoning why this is the case. After Sending the last prompt (Nr. 1400) and receiving the ``END OF Terahertz Images" Tag, please output the findings based on the $<$ FINAL Return Format $>$ and $<$Warnings$>$.
\\\\ 
- Prompt Format Each Image
\\
$<$Sending the next Image now. Analyse it and answer with reasoning. Here is the Image Nr. [Image NR]$>$ 
\\\\\\  
$<$BEGIN Return Format Each Image$>$\\
$\#\#\#$ Frame Number\\
$\#\#\#$ Description: Give a clear description of the region in the image where C4 appears.\\
$\#\#\#$ Classification: Give a clear classification of either "Yes C4" or "No C4".\\
$\#\#\#$ Memory Update if significant changes detected.\\
$<$END Return Format Each Image$>$ 
\\\\ 
$<$BEGIN FINAL Return Format$>$\\ 
Most confident frame index (frequency) where C4 is visible.\\  
A clear description of the region in that frame where C4 appears.\\ 
List of frames (frequency indices) and approximate positions where C4 is confidently detected.\\ 
$<$END FINAL Return Format$>$ 
\\\\
$<$BEGIN Warnings$>$\\ 
Do not use frames 0–50 and 1350–1399 unless there are strong, obvious indicators.\\
No raw THz signal processing or additional pre-processing is needed.\\
Only visual and logical analysis of the provided plots is required.\\
Analyze all frames at a time and keep memory of previous ones.\\
Focus analysis on the central 1000 frames (250–1250).\\
Some noise may be present.\\
$<$END Warnings$>$ 
\\\\
- Context Dump\\ 
Terahertz (THz) images are captured using electromagnetic waves in the terahertz band (roughly 0.1–10 THz). In practice, a THz imaging system illuminates the object with THz radiation and records the reflected or transmitted signal. Commonly, coherent pulsed THz systems (THz time-domain spectroscopy) are used: an ultrafast laser generates short THz pulses, and a detector measures the time-domain waveform. This yields both amplitude (intensity) and phase information at each pixel, allowing reconstruction of 2D images or tomographic 3D slices. THz imaging is non-ionizing and can see through many non-metallic materials, providing contrast based on dielectric properties.
\\\\ 
In our THz images, intensity refers to the power of the detected terahertz (THz) field at each pixel location. It is computed as the square of the magnitude of the complex THz field, reflecting the energy content at that point in space. Before visualization, the raw complex-valued data is preprocessed to ensure consistent orientation and structure. This includes operations such as magnitude calculation, squaring, and coordinate correction. To enhance contrast and interpretability, the intensity values are then scaled using softmax normalization, which maps them into a relative range - typically [0, 1]—based on their exponential distribution. The final intensity is expressed in arbitrary units, representing relative differences in THz power absorption or reflection. When interpreting intensity, brighter (higher intensity) regions generally correspond to areas where more THz power is transmitted or reflected, indicating lower absorption by the material. Conversely, darker (lower intensity) regions suggest higher absorption or scattering, which may indicate structural boundaries, material differences, or hidden features. The intensity map thus serves as a spatial representation of the material's electromagnetic response to THz radiation.
\\\\
Phase in THz imaging refers to the angle of the complex electromagnetic wave at each pixel in the image. It indicates the timing or delay of the wave as it propagates through or reflects off the sample. To calculate the phase, the raw data is first processed by applying a windowing function to the complex field data, followed by a Fourier Transform to shift the frequency domain. The phase is then extracted from the complex field at a specific depth, representing the temporal shift or time delay at each point in the sample. When interpreting the phase image, shifts in phase can reveal variations in material properties, such as the refractive index and thickness of the sample. Phase contrast highlights subtle structural details that might not be visible in the intensity image, making it useful for identifying internal material differences or features hidden beneath the surface.
\\\\
The THz images were acquired using a reflection-mode Terahertz Time-Domain Imaging (THz-TDI) setup with a single transmitter and receiver. A 2D raster scan was performed across the sample surface, capturing time-resolved THz signals at each (x, y) position. A lens was used to focus the THz beam, and the time-domain data were converted into depth-resolved images via a Fourier Transform (FFT), enabling the formation of a 3D image volume. The scanning produced 1400 depth samples (z-axis) per lateral position, providing fine resolution along depth. The lateral sampling resolution was 0.2625 mm, which is considered high and allows for detailed spatial mapping.
\\\\
WHAT TO EXPECT\\ 
In Terahertz (THz) imaging, both geometry and material response influence the output. The dataset includes two distinct target types: (1) A metal plate (2) A metal-coated object, possibly containing C4 explosive material underneath.These are visualized through two types of images per depth layer: (1) Intensity images (with colormap viridis), showing reflected signal strength normalized via softmax to the range [0, 1]. (2) Phase images (with colormap twilight), showing the wave propagation delay from $-\pi$ to $\pi$ radians. One frame alone cannot give one this information. It is essential for reasonable deduction to consider change in intensity and phase across multiple frames as the depth increases and frequencies change. Expected Observations:\\\\
1.	Metal\\
Intensity (viridis):\\ 
The metal plate should appear as a bright region with fairly uniform intensity due to strong, flat-surface reflection. Edges may appear slightly brighter due to specular highlights or slight angle differences. This observation should remain more-or-less consistent across depth frames as frequencies change.
\\\\
Phase (twilight):\\ 
The phase map should reveal clear boundaries of the metal, possibly forming sharp phase discontinuities at boundaries. Inside the metallic region, the phase may be relatively uniform or show subtle gradients depending on plate thickness or surface tilt. This observation should remain more-or-less consistent across depth frames as frequencies change. 
\\\\\\
2. C4\\
Intensity (viridis):\\
This object should have non-uniform reflections that stand out from the background. One giveaway would be substantial changes in intensities from one frame to another. 
\\\\
Phase (twilight):\\
This object should have non-uniform reflections that stand out from the background. One giveaway would be substantial changes in intensities from one frame to another. 
\\\\
Please note that the C4 might be inside metallic covering and thus the region that contains C4 might sometimes mimic the expected properties of metal, especially in the initial and the last frames from the 1400 frames.  
\\\\
DESCRIBE IMAGES (Intensity $\&$ Phase)\\
Each Terahertz image represents a 2D scan at a specific frequency, indicated at the top of the image. The image is divided into two main components:\\\\
Left Panel - Intensity: Displays the normalized reflection intensity of the signal, using the viridis colormap. The values are softmax-normalized and scaled to the range [0,1], highlighting relative signal strength across the scanned area.\\\\
Right Panel - Phase: Shows the phase information of the reflected wave, using the twilight colormap. Values range from [-$\pi$,$\pi$] radians, representing the wave propagation delay, which is sensitive to material properties and geometric features.\\\\
Each panel includes X and Y axes that indicate the spatial dimensions of the scan, allowing for interpretation of physical positioning within the imaged area. ''}
\footnotesize\texttt{(1 system prompt.txt)}
\end{quote}

\subsection{Prompts Used by Classification Type}
This section summarizes which prompts were employed depending on the classification setting.
\\\\
\textbf{Zero-shot Classification} uses the following prompts:
\begin{itemize}
    \item Task Instruction
    \item Query\\
\end{itemize}
\textbf{One-shot Classification} uses the following prompts:
\begin{itemize}
    \item Task Instruction
    \item Demonstration Set
    \item Query
\end{itemize}

\section{THz Signal To Frequency Spectra Plots}
\begin{algorithmic}[1]
\Statex \hrulefill

\Statex \textbf{Input:} Raw THz data from .mat file (3D tensor of shape \texttt{(NF, NX, NY)})
\Statex \textbf{Output:} Intensity and Phase spectra plots

\Statex \hrulefill

\Statex \textbf{Load and reconstruct Data:}
\Statex
\State \textit{Extract metadata (frequency sampling, number of slices)}
\State \textit{Reconstruct 3D complex-valued data tensor from segmented blocks}
\State \textit{Convert to PyTorch tensor:} \texttt{data\_complex[NF, NX, NY]}

\Statex \hrulefill
\Statex \textbf{Apply Signal Processing Pipeline:}
\Statex
\State \textit{Apply 1D Hamming window along frequency axis (NF)}
\State \textit{Apply FFT along frequency axis:} \texttt{data\_fft = FFT(data\_complex, dim=0)}
\State \textit{Shift zero-frequency component to center:} \texttt{data\_fft = fftshift(data\_fft, dim=0)}

\Statex \hrulefill
\Statex \textbf{Compute Features at Depth Layer $d$:}
\Statex
\Statex $\#$\textsc{Define $I$ as intensity, $\phi$ as phase}
\State $I \gets$ \textit{Intensity image (magnitude squared)}
\State $\phi \gets$ \textit{Phase image (angle of complex values)}
\Statex
\Statex $\#$\textsc{Extract complex slice at target depth:}
\State $D \gets \texttt{processed\_data}[d, \dots]$
\Statex
\Statex $\#$\textsc{Compute Intensity as squared magnitude:}
\State $I \gets |D|^2$
\Statex
\Statex $\#$\textsc{Compute Phase as angle of complex values:}
\State $\phi \gets \arg(D)$
\Statex
\Statex $\#$\textsc{Flip vertically and transpose for orientation:}
\State $I \gets \texttt{transpose}(\texttt{flipud}(I))$
\State $\phi \gets \texttt{transpose}(\texttt{flipud}(\phi))$
\Statex
\Statex $\#$\textsc{Convert to NumPy format:}
\State $I \gets I.\texttt{cpu().numpy()}$
\State $\phi \gets \phi.\texttt{cpu().numpy()}$
\Statex \hrulefill
\Statex \textbf{Normalize and Visualize:}
\Statex
\Statex $\#$\textsc{Normalize intensity for visualization}
\State Generate and output 2D intensity and phase plots for all spatial positions $(NX, NY)$.
\end{algorithmic}

\section{Crop Used For In-Context Learning}
\label{sec:appendix:crop_used_ICL}
\input{latex_for_figures/THz-Data-Example-Crop}
\noindent The crop used for the one-shot ICL is shown by \Cref{fig:THz-Data-Example-Crop}.

\section{Metrics Used}
\label{sec:appendix:metrics_used}
To evaluate the effectiveness of model predictions and the impact of in-context learning (ICL), we use a combination of standard classification metrics and a custom metric for analyzing prediction shifts. These provide both quantitative performance insights and a finer-grained understanding of prediction dynamics.
\subsection{Standard Classification Metrics}
We evaluate model predictions against human-annotated binary labels using four standard classification metrics: Accuracy, Precision, Recall, and F1-score. 
These metrics assess both overall correctness and the balance between false positives and false negatives.

\noindent\paragraph{Accuracy}  
Accuracy measures the proportion of total predictions that are correct. It gives a general indication of how often the model is right across all classes.
\begin{equation}
\text{Accuracy} = \frac{TP + TN}{TP + TN + FP + FN}
\end{equation}
where \( TP \) is the number of true positives, \( TN \) is true negatives, \( FP \) is false positives, and \( FN \) is false negatives.

\noindent\paragraph{Precision}  
Precision measures the proportion of positive predictions that are actually correct. In this context, it reflects how often the model correctly identifies the presence of C4 when it predicts “C4 present.”
\begin{equation}
\text{Precision} = \frac{TP}{TP + FP}
\end{equation}

\noindent\paragraph{Recall}  
Recall (also known as sensitivity) quantifies the proportion of actual positives correctly identified. It indicates how effectively the model detects C4 whenever it is present.
\begin{equation}
\text{Recall} = \frac{TP}{TP + FN}
\end{equation}

\noindent\paragraph{F1-score}  
The F1-score is the harmonic mean of precision and recall, balancing both metrics in a single value. It is especially useful when dealing with class imbalance or when both false positives and false negatives carry significant consequences.
\begin{equation}
\text{F1-score} = \frac{2 \cdot \text{Precision} \cdot \text{Recall}}{\text{Precision} + \text{Recall}}
\end{equation}

\subsection{Prediction Changes Metric}
To better understand the direct impact of In-Context Learning, we introduce a complementary evaluation method that tracks prediction changes when moving from zero-shot to one-shot settings. As shown in Table~\ref{tab:experiment-result-impact-ICL}, each model prediction is classified into one of four mutually exclusive categories:\\
\begin{itemize}
    \item \textbf{Improvement}: Cases where the model predicted incorrectly in the zero-shot setting but predicted correctly after ICL was applied.\\
    \item \textbf{Decline}: Cases where the model predicted correctly in the zero-shot setting but predicted incorrectly after ICL was applied.\\
    \item \textbf{No Improvement}: Cases where the model predicted incorrectly in the zero-shot setting and remained incorrect after ICL was applied.\\
    \item \textbf{No Decline}: Cases where the model predicted correctly in the zero-shot setting and remained correct after ICL was applied.\\
\end{itemize}
This metric enables a concise, directional analysis of ICL effects at the level of individual predictions. It highlights when and how the demonstration set affects classification outcomes, both positively and negatively. An ideal ICL-enhanced model would maximize the number of \textit{Improvement} and \textit{No Decline} cases, while minimizing \textit{Decline} and \textit{No Improvement} outcomes.

\section{Limitations}

While this work marks a significant step toward leveraging Vision-Language Models (VLMs) for security-related THz imaging tasks, several limitations remain. First, despite the promise of zero-shot VLMs, their standalone performance remains limited in reliability and precision for high-stakes applications \cite{phan2025humanity_last_exam}. Our results show that one-shot In-Context Learning (ICL) can offer modest improvements, but the system still lacks robustness across all cases.

Furthermore, although this study focuses on open-weight models for accessibility and reproducibility, proprietary models with larger context windows or dedicated fine-tuning may offer superior performance. Exploring such approaches is an important direction for future work.

Another challenge lies in the quality of the input data. The THz frequency spectra used in this study contain aliasing artifacts, which may hinder the model’s ability to extract semantically meaningful features. Addressing these issues, e.g., through signal-domain preprocessing or aliasing-aware design strategies \cite{agnihotri2024beware}, could enhance downstream classification accuracy.

Overall, while our framework lays the foundation for using interpretable VLMs in THz-based decision support systems, substantial progress is needed before real-world deployment is feasible.

%% file: latex_for_figures/THz-Data-Example-Crop.tex
\begin{figure}[htbp]
    \centering
    \includegraphics[width=\linewidth]{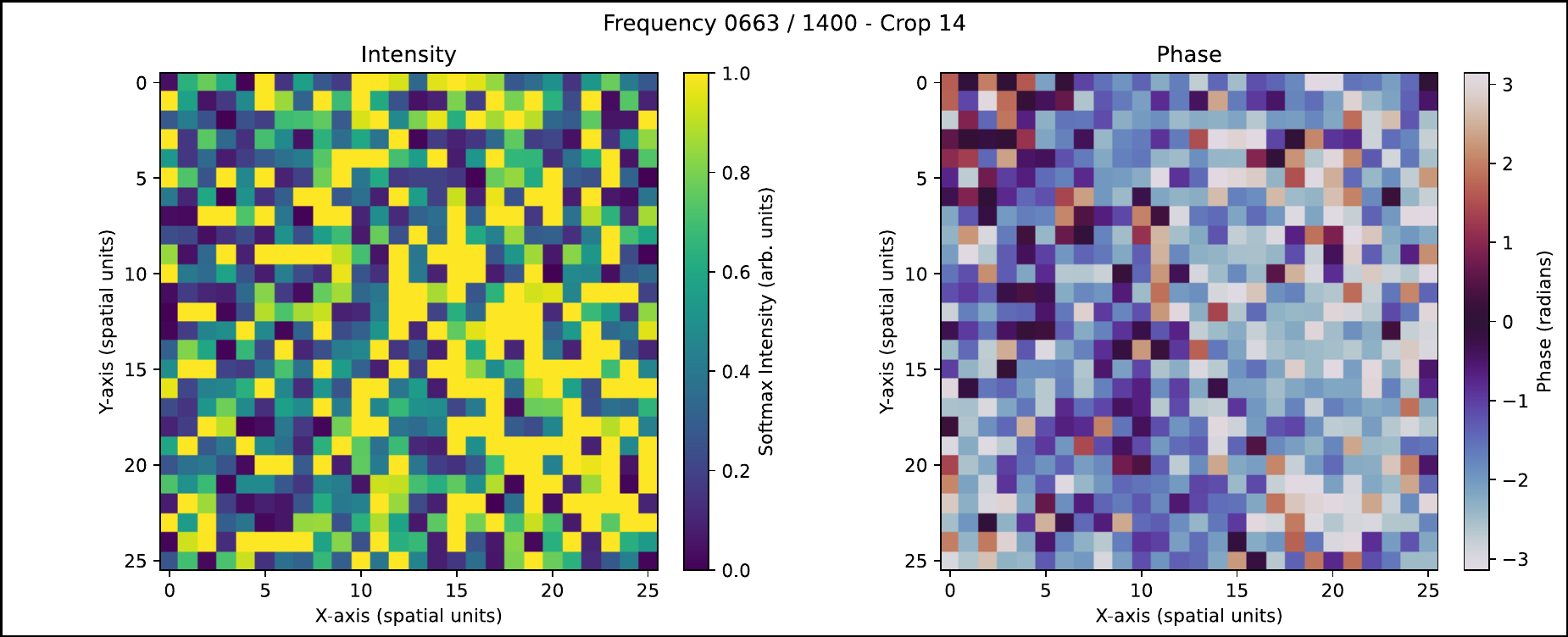}
    \caption{Example crop extracted from one of the 1,400 frames in the dataset. This crop comes from Frame Number 0663 (the same frame seen in Figure \ref{fig:THz-Data-Example}), which displays its intensity and phase plots with the visible “C4” circle. The left side shows the intensity plot, while the right side shows the corresponding phase plot. This crop corresponds to the area shown in Figure \ref{fig:THz-Data-Image-To-Crop}.}
    \label{fig:THz-Data-Example-Crop}
\end{figure}